\documentclass[runningheads]{llncs}
\usepackage{graphicx}
\usepackage{geometry}
\usepackage{booktabs}
\usepackage{multirow}
\usepackage{cite}
\usepackage{amsmath}

\begin{document}
\newgeometry{margin=1.6in}

\title{A Human-Robot Mutual Learning System with Affect-Grounded Language Acquisition and Differential Outcomes Training
\thanks{All authors declare that they have no conflicts of interest. Supported by an EUTOPIA Undergraduate Research Support Scheme (EURSS) grant to AM and SS for a research stay at ETIS, and by Brain +. The work was carried out partly at ETIS, CY Cergy Paris University, and partly at the Dept. of Applied IT, University of Gothenburg.}}
\titlerunning{Human Robot Mutual Learning System}
%
\author{Alva Markelius\inst{1, 2}\orcidID{0009-0003-4580-9997}\thanks{Corresponding author's email: ajkm4@cam.ac.uk} \and
Sofia Sjöberg\inst{2}
\and
Zakaria Lemhauori\inst{3,4}\orcidID{0009-0008-0235-8897}
\and
Laura Cohen\inst{3}
\and
Martin Bergström\inst{2}
\and
Robert Lowe\inst{2}\orcidID{0000-0002-0307-3171}
\and
Lola Ca\~{n}amero\inst{3}\orcidID{0000-0001-6903-1348}
}
\authorrunning{A. Markelius et al.}
%
\institute{University of Cambridge, Department of Philosophy, Sidgwick Ave, Cambridge, UK \and
DICE lab, University of Gothenburg, Department of Applied IT, Forskningsgången 6, Gothenburg, Sweden
\and
ETIS Lab, CY Cergy Paris University - ENSEA - CNRS UMR8051, France
\and
Artificial Intelligence Lab, Vrije Universiteit Brussel (VUB), Pleinlaan 9, Brussels, Belgium
\\
}
\maketitle  

\begin{abstract}
This paper presents a novel human-robot interaction setup for robot and human learning of symbolic language for identifying robot homeostatic needs. The robot and human learn to use and respond to the same language symbols that convey homeostatic needs and the stimuli that satisfy the homeostatic needs, respectively. We adopted a differential outcomes training (DOT) protocol whereby the robot provides feedback specific (differential) to its internal needs (e.g. `hunger') when satisfied by the correct stimulus (e.g. cookie). We found evidence that DOT can enhance the human's learning efficiency, which in turn enables more efficient robot language acquisition. The robot used in the study has a vocabulary similar to that of a human infant in the linguistic ``babbling'' phase. The robot software architecture is built upon a model for affect-grounded language acquisition where the robot associates vocabulary with internal needs (hunger, thirst, curiosity) through interactions with the human. The paper presents the results of an initial pilot study conducted with the interactive setup, which reveal that the robot's language acquisition achieves higher convergence rate in the DOT condition compared to the non-DOT control condition. Additionally, participants reported positive affective experiences, feeling of being in control, and an empathetic connection with the robot. This mutual learning (teacher-student learning) approach offers a potential contribution of facilitating cognitive interventions with DOT (e.g. for people with dementia) through increased therapy adherence as a result of engaging humans more in training tasks by taking an active teaching-learning role. The homeostatic motivational grounding of the robot's language acquisition has potential to contribute to more ecologically valid and social (collaborative/nurturing) interactions with robots.

\end{abstract}

\newgeometry{margin=1.08in}

\section{Introduction}
The increased implementation of socially assistive robots (SAR) to provide motivational, social, pedagogical, and therapeutic assistance raises the questions of how components such as emotion expression, language and motivation can be grounded in homeostatic interaction and collaboration with environment and other social agents \cite{lowe2016grounding, mataric2016socially}. This is particularly relevant in order for SAR to act appropriately and to enable e.g. formation of affective social bonds \cite{khan2022long}, adaptive acting upon and expressing emotion \cite{lowe2016grounding} and behaviour grounded in motivation, physiological needs and environmental conditions \cite{cos2013hedonic, lewis2016hedonic}. These components are important to facilitate interactions e.g., for vulnerable populations, which constitute one of the main target groups for SAR \cite{mataric2016socially}, including elderly and children in the contexts of learning and cognitive interventions. Human-robot collaboration and interaction setups with SAR are gaining traction as a promising approach for facilitating and mitigating cognitive interventions \cite{alnajjar2019emerging}. This includes memory training for people with e.g. mild cognitive impairment (MCI), which is a clinical phase between normal aging and dementia \cite{hampel2016rising} or dementia \cite{aguera2015social, andriella2020cognitive, kim2015structural}. Cognitive interventions have emerged as crucial alternatives to pharmacological treatments as they provide accessible, effective, and scalable solutions to address issues that often have limited or no available pharmacological alternatives \cite{alzheimer_society_2022}. In particular, for patients with MCI, cognitive training is gaining attention as a promising form of intervention because of its effectiveness in slowing down, or even halting the advance to dementia \cite{hong2015efficacy}. However, treatment adherence in cognitive interventions tends to be low, in particular for older adults, resulting in discontinued or uncompleted treatments \cite{fernandez2019adherence, walsh2019association}. This issue is likely the result of lack of engagement in the treatments, both on an affective level and as a result of boredom, lack of social accountability and means to consistently carry out the intervention.  

Previous research has included SAR to facilitate engagement in cognitive interventions, where usually the robot takes a supporting and teaching role \cite{kim2015structural}. We present a collaborative human-robot interactive setup that includes a mutual learning system for both robot and human, offering a novel contribution to implementing SAR for cognitive interventions. Mutual learning has great potential to facilitate both the advancement of developmental robotics as well as memory training in humans. We define it here as teacher-student collaborative learning, where both robot and human act as teacher and student simultaneously, such as in the learning paradigm described in \cite{eSwapp_2021}. The setup allows the human to take a more active role as a caregiver to a robot learning to acquire language through homeostatic social and affective interaction. As such, this approach is expected to foster greater engagement, a heightened sense of control, and a deeper feeling of contribution than previous approaches where the human has a more passive role. Additionally, the present study includes differential outcomes training (DOT), as a methodology to enhance learning for the human and subsequently increase robot learning efficiency (as described in the following sections). This paper accounts for the methodology of the present collaborative setup and presents the results from an initial pilot study conducted to gain insights into the social and affective experience of the humans and its effectiveness for the robot software architecture in this specific context. This is the first time to the authors knowledge that a human-robot mutual learning methodology has been developed that includes specialised training protocols, e.g. DOT for the human and homeostatic language acquisition for the robot in combination to facilitate the learning of both. Additionally, as DOT is a training protocol specifically suited for neurodegenerative disorders \cite{plaza2012improving, vivas2018enhancement}, this methodology offers a novel way for the human to be more proactive in the training, as compared to many other cognitive training interventions. The rest of the paper is set out as follows: Section 2 accounts for previous related research; Section 3 provides an overview of the methodological approach in the present study; Sections 4 and 5 account for pilot study results and discussion, respectively.

\section{Related work}
The work presented in this paper lies in the intersection of developmental robotics (including language acquisition, motivation and affect development) and research into the use of homeostatic mechanisms in robotics (including emotion and social grounding of interactive robot activity). Finally, one of the main components of this work includes memory training of the human part of the interactive scenario, which builds on previous work in cognitive science and psychology (differential outcomes training and cognitive interventions).

\subsection{Homeostatic Developmental Robotics and Language}
This study uses and builds upon a previously developed human/robot interaction setup developed by Lemhaouri et.al. \cite{lemhaouri2022role}. The original setup concerns language acquisition of a humanoid robot through interaction with a human caregiver and is grounded in affective motivation as the main driving force behind language acquisition. In this setup, the robot possesses a vocabulary similar to that of an 8-month-old infant during the linguistic babbling phase \cite{davis1994organization}. The setup is based on trial-and-error, where the robot expresses an internal need and the caregiver has to learn this need by probing, i.e. providing the robot with an object to fulfill the need. If guessed correctly, the word (expressed need) will be reinforced to be associated with the particular need. Previous homeostasis based work in robotics has focused on biologically inspired approaches to facilitate interaction between robots and humans and robot inner motivations as a driving factor to foster decision making, behavior and emotion in relation to environment and other social agents \cite{cos2013hedonic, yang2017companion}. For instance, dynamic, biologically-inspired social mechanisms in artificial agents have been investigated for forming long-term affective bonds \cite{khan2022long} and social learning \cite{bartoli2020mechanisms} as well as more complex mechanisms such as reasoning and interacting on perceived affective and intentional state \cite{pieters2017human}. This has previously been shown to be relevant to robots deployed in elderly care \cite{yang2017companion} where the inner motivations of the robot enable it to assist, converse and care depending on environmental and social stimuli. This functional approach differs from many other computational approaches of robot language development as it acknowledges the role of motivation, affect and semantic meaning in linguistic acquisition \cite{lemhaouri2022role}. Section 3.1 provides a detailed overview of the methodology and the robot software architecture.

\subsection{Differential Outcomes Training}
DOT refers to the learning of unique stimulus-response pairs when certain stimuli are associated with certain rewards/outcomes. More specifically, DOT entails correct responses on a given task being reinforced by feedback that is unique to both the response and stimulus. For example, the response of pushing forward a "cookie" object (response) following the stimulus (utterance ``hungry'' in an unknown language) would yield the rewarding outcome of a specific social expression (e.g. smile). For different stimulus-response pairings (with a correct response) another unique outcome would be expressed (e.g. head nod). In non-differential outcomes tasks, this feedback following the correct response would be randomized (either smile or head nod irrespective of stimulus-response). The DOT approach, contrasting with standard reward-based training where the same reward is given irrespective of stimulus and response `pairings', has consistently been shown to enhance discriminative learning and memory performance accuracy in animals, children and adults \cite{esteban2014differential, esteban2015spatial, peterson1980effects}. Furthermore, evidence suggests it is effective in elderly people with memory impairments such as Alzheimer's disease \cite{plaza2012improving, vivas2018enhancement}. A recent systematic review with meta-analysis \cite{mccormack2019quantifying} revealed that the impact of DOT on learning and memory exhibits a medium to large effect size as compared to control condition comparisons. This review also found that the positive learning effect of DOT was found in several studies deploying persons with dementia or MCI. The use of repeated sessions of training (i.e. for full cognitive interventions) may, thereby, be worth exploring as a means for mitigating the symptoms of early stage dementia or MCI. Our previous work has suggested using DOT in combination with SAR \cite{markelius} with promising results, showing that the DOT effect can also be obtained in the context of human-robot interaction. DOT-effects have been found in human-human interactive learning \cite{rittmo2020} and also theorized in the context of human-robot interactive learning using facial expressions and perceived outcomes of others \cite{lowe2019vicarious}. Furthermore, differential outcomes expectancies have been computationally modelled in the context of learning and interaction using reinforcement learning \cite{lowe2016minimalist}.
In this study, the DOT is presented in the interaction between the human and robot, as the feedback provided by the robot is differential to certain homeostatic motivations and needs (based on reinforcement learning). Furthermore, the nature of the interaction entails a teacher-student relationship, whereby the human and robot serve as both teacher and student to each other (mutual learning). Section 3.2 provides a more detailed overview of how DOT is implemented in this setup.

\begin{figure}
    \centering
\includegraphics[width=0.5\textwidth]{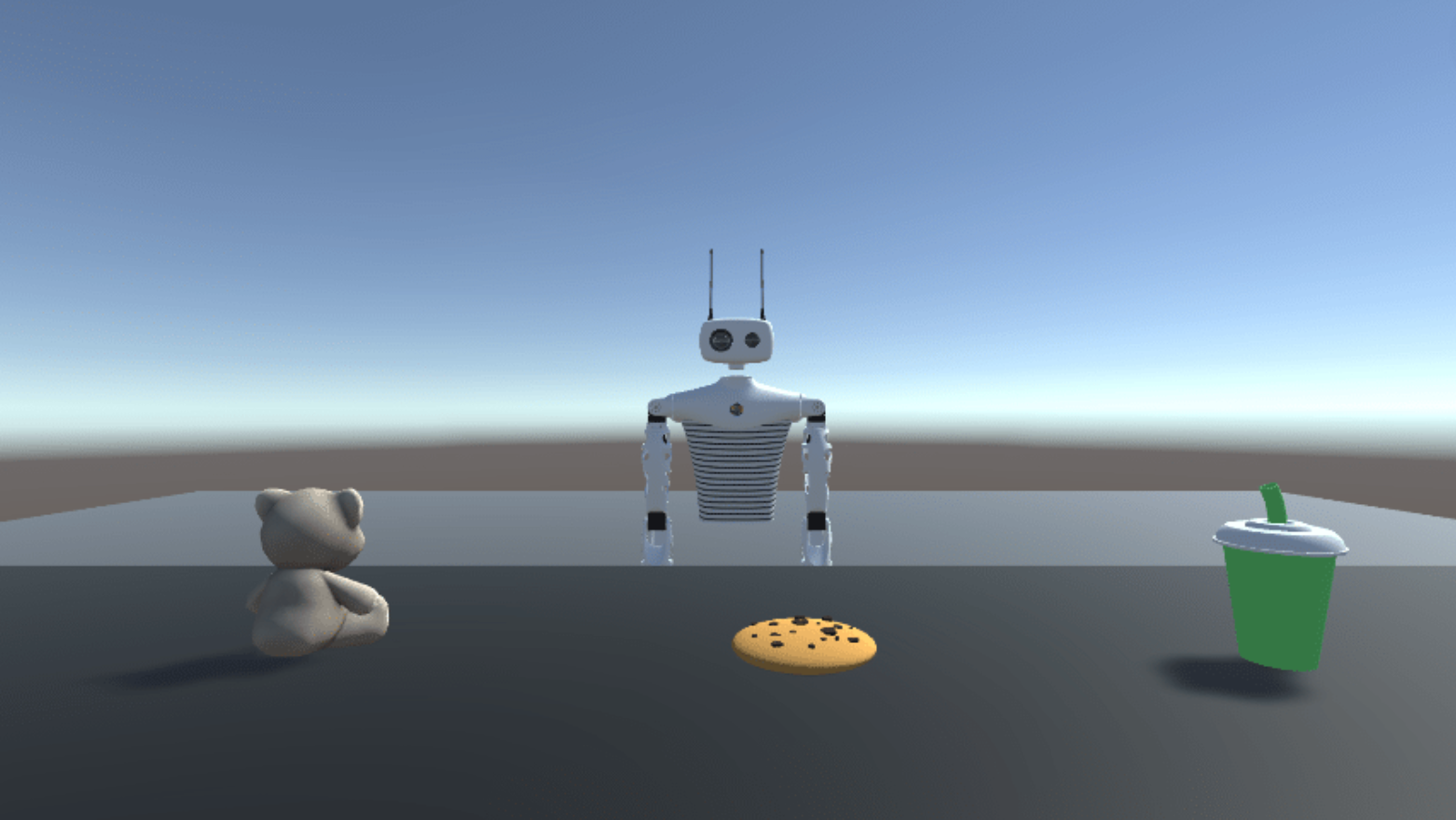}
\caption{The robot as seen by the participants in the study (human caregivers). The visual field includes the three objects that can be given to the robot to fulfill its internal needs} \label{fig1}
\end{figure}

\section{Methodology}
The robot used in this study is the simulated (Unity implemented, version 2020.3.11f1) SDK-version of the Reachy robot (version SDK 0.5.1), developed by Pollen Robotics, a humanoid robot able to expressively move its head, antennae and arms (See Figure 1). The simulated setup closely reflects the dynamics and functions of a real world version of the setup, using the physical version of the Reachy robot (which has previously been tested and piloted). Thus includes the software architecture, the roles of the human and robot and the setup design and procedure.

\subsection{Robot Software Architecture}
The robot software architecture is built upon \cite{lemhaouri2022role} who developed a model for affect grounded language acquisition, which is based on creating associations between vocabulary and internal needs through human `caregiver' interaction. The setup includes a reinforcement learning cycle (using a multi-armed bandit algorithm) where the robot verbally expresses an internal need as determined by affective motivation by requesting an object from the caregiver. There are three different internal needs (hunger, thirst and curiosity) and three respective objects that can fulfill those needs (cookie, drink, teddy bear). Initially, the robot generates words randomly when one need becomes more significant than the others, the caregiver being unaware of which need the robot is expressing. Upon hearing the robot's vocalization, the caregiver selects an object and hands it to the robot. If the provided object satisfies the robot's need, the motivation associated with that need diminishes, and the robot receives a reward of +1, expressing its satisfaction through giving positive audiovisual feedback. However, if the object does not fulfill need, the word is penalized with a reward of -1, reducing the probability of using that word in a similar context, and the robot gives negative audiovisual feedback \cite{lemhaouri2022role}. The robot software learning model consisted originally of three modules, the motivation, visual perception, and phonological modules \cite{lemhaouri2022role}. In the present study, a fourth module was added, the DOT feedback module. 

\subsubsection{Motivation Module:}
The three internal needs (hunger, thirst, curiosity) are determined by a homeostatic variable that continuously decreases over time until the need is fulfilled, which makes it increase. This can be compared to blood glucose level that decreases over time and increases after a meal. The robot drive $d_i(t)$, representing the urge to act and satisfy the need $i$, is defined as the difference between the current homeostatic variable and its optimal value. The robot’s motivation to satisfy a need depends on the related drive and the intensity of the stimulus that can satisfy it\cite{avila2004using}:
\begin{equation}
\label{eq:1}
    m_i(t) = d_i(t) + d_i(t).s_i
\end{equation}
$s_i$ is the intensity of the stimulus estimated by the visual perception module (described below). The robot’s motivation to act on a certain internal need is determined by a threshold that, if it is reached for a certain need before another, leads the robot to express a `will' to satisfy that need (babble) based on the phonological module. The need is satisfied based on the visual perception module (if the robot is given the right object to satisfy the need or not).  
\subsubsection{Visual Perception Module:} 
This module helps modulate the motivation system and determines if a need is satisfied based on the robot’s ability to recognise objects given by the caregiver (cookie, drink, teddy bear) in response to its expression of an internal need.  This gives the robot the ability to identify and label objects within its visual environment, while also visually associating the internal needs that the recognised objects can satisfy. The model used for object detection and recognition is a combination of a modified version of Kohonen's self-organizing map clustering algorithm\cite{kohonen2012self} and a multi-class perceptron in which the weights update $\Delta \omega$ follows the Widrow-Hoff learning rule:
\begin{equation}
\label{eq:2}
\Delta \omega_{ij}=\epsilon VF_i(RIS_j-ISP_j)
\end{equation}
With :\\
VF: The visual features of the object calculated by SIFT algorithm\cite{lowe1999object}.\\
$\epsilon$: the learning rate.\\
RIS: the robot internal state.\\
ISP: internal state prediction.\\

\subsubsection{Phonological Module:}  
This module allows the robot to verbally express its internal needs by a text-to-speech functionality. The robot's language repertoire comprises two-syllable words which are chosen from some of the most frequent syllables commonly used by an 8-month-old human infant (examples include ``nana”, ``wada”, ``pada”). The robot initially starts by randomly selecting a word whenever one particular internal need becomes more pronounced than the others. As the robot receives an object from the caregiver it learns to associate certain words in its vocabulary with certain internal needs.

\subsubsection{DOT Feedback Module:}
In addition to the three above mentioned modules, the current study included a DOT feedback module. This module determines the type of positive feedback given for a certain internal need. If the robot’s need is not fulfilled the robot always expresses the same negative audiovisual feedback (looking down, lowering antenna and making a sad sound). However, if the need is fulfilled, the robot expresses a certain positive feedback associated with each internal need. Each need had different happy beeping sounds and different movements (curiosity: wagging antennae, hunger: happy arm movement, thirst: nodding head movement). In other words, each internal need has its own sound and movement associated with it, which the robot consistently expresses when that need is satisfied. Each feedback response is therefore differential with respect to each internal need fulfillment. In the non-DOT condition the feedback included the same three movements but they were instead randomised for internal needs (i.e. not tied to specific needs). For a comparison of how the interactive scenario differs between the DOT and non-DOT based feedback see Figure 2. The movements were implemented as inspired by how animals tend to express contentment or happiness (such as dogs wagging their tail). Tails and wagging motions have previously been implemented with robots to express happiness \cite{singh2013dog}, and in the present study we implemented a similar wagging motion on both the robot's antennae and arms. 
\begin{figure}
\includegraphics[width=\textwidth]{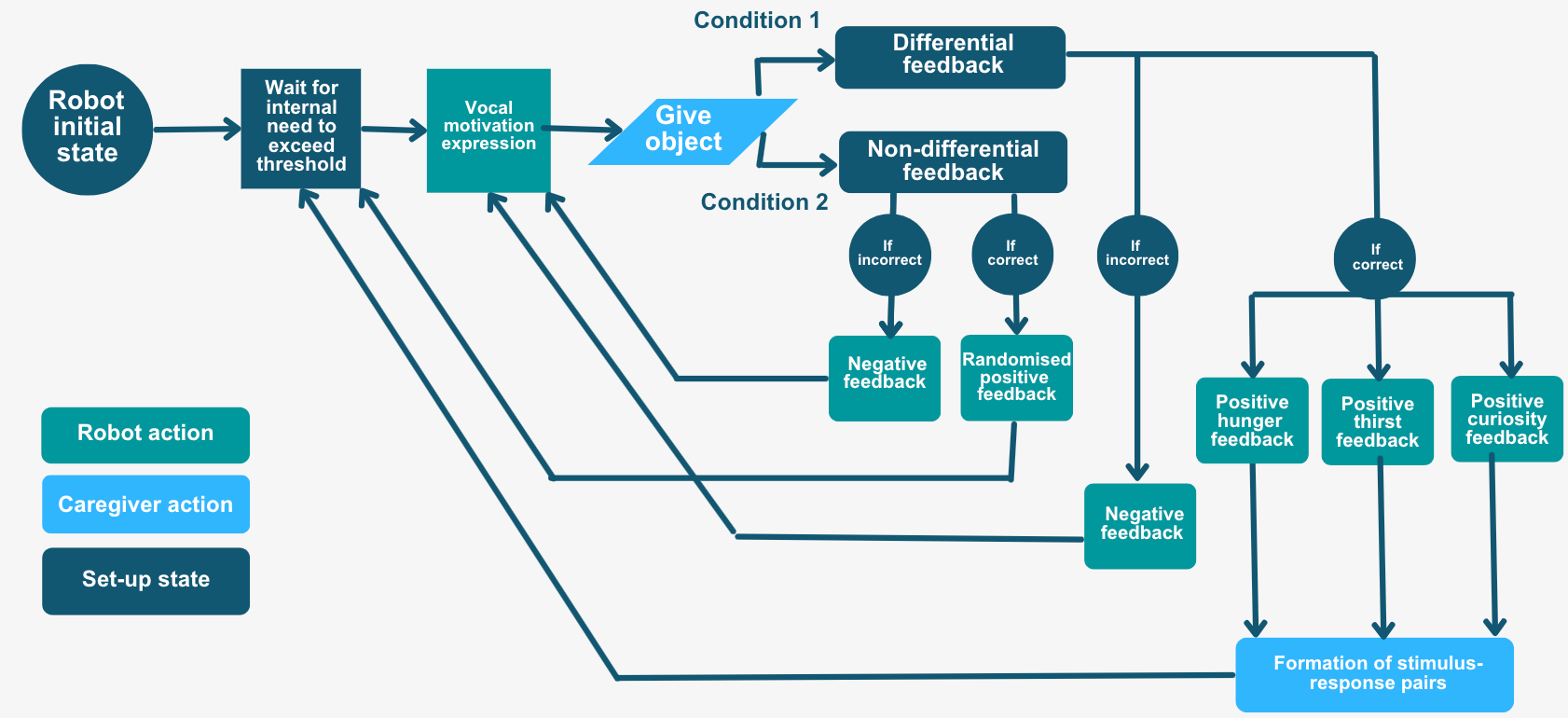}
\caption{Flowchart of one iteration of the experimental procedure, depending on either DOT or non-DOT conditions. The experiment consisted of 12-16 iterations depending on the fulfillment of the robot needs.} \label{fig1}
\end{figure}

\subsection{Experimental Setup}
The setup was tested in an initial pilot study to try out the functionality of the robot architecture including the new DOT module as well as getting some first insights into participants’ experience of interacting with the robot in this particular setup. The pilot study had only one independent variable (two levels): with or without the DOT feedback module (condition 1: DOT, condition 2: non-DOT). Furthermore, the pilot study solely included a virtual version of the robot and the interaction was thus carried out through a laptop interface. For a visualisation of the robot as seen by the participants see Fig 1. Six healthy Swedish adults (not suffering from cognitive impairments or memory deficiencies) participated in the pilot study, aged between 24 and 37 years (M= 29). Half of the participants were in the DOT condition and half in non-DOT condition. Informed consent was ensured by all of the volunteering participants by provision of informed written consent forms. The study was conducted in accordance with the WMA Declaration of Helsinki. The participants did not receive any compensation. 

\subsubsection{Procedure}
The experiment was carried out according to the following procedure. After having been given instructions and information about the study the participant sat down in front of the laptop with the Unity virtual version of the robot setup (as seen in Fig 1). They were instructed to give the robot an object when it expressed an internal need, and to remember what object was associated with what word. The experiment lasted for approximately 13-16 iterations which took about 10 minutes, iterations ended automatically as a function of learning. For a full flowchart of one experimental iteration see Fig 2. After the interaction with the robot, the participants filled in a survey about their experience and affective state associated with the interaction. Finally, a semi-structured interview was conducted with each participant about their experience and perception of the interactive scenario and the robot.

\subsubsection{Data Collection}
The robot's ability to learn to associate certain words in its vocabulary with a certain internal need is determined by convergence rate. Full convergence is reached when the robot achieves a state where it consistently selects appropriate words based on its internal state. Additionally, this enables the caregiver to understand the robot, facilitating the robot's ability to obtain the desired objects. The evaluation metric used to determine convergence rate is the moving average of rewards (MAR) received by the robot:
\begin{equation}
MAR_n=\begin{cases}
\displaystyle\frac{1}{n} \sum_{i=1}^n R_i
& \text{if } n<m \\\\
\displaystyle\frac{1}{m} \sum_{i=n-m+1}^n R_i  & \text{Otherwise}\\ 
\end{cases}
\end{equation}
$m$ is the number of previous reward values used to calculate the MAR at the iteration $n$. In this experiment we chose $m$=5. The convergence time, on the other hand, is defined as the number of iterations required for the robot to reach 90\% convergence. Thus, the dependent variable in terms of robot language acquisition is average convergence rate. For determining the participants' affective experience of the setup self-report data was collected using the dimensional Self Assessment Manikin (SAM) Scale \cite{bradley_lang_1994}. It was used with 3 dimensions, valence (happy/unhappy), arousal (stressed/not stressed) and dominance (in control/not in control) each to be rated between 1 to 5. To complement the scale, the survey also contained likert-scale based questions about the specific setup. Finally, the semi-structured interviews were recorded and transcribed to enable a thematic analysis as well as case by case qualitative analysis. The thematic analysis was conducted by two of the authors individually to encompass a more comprehensive and objective identification of themes and key words. Initially, general code words were identified in the transcribed interviews, and were then categorised into more general themes. The most commonly occurring themes, mostly relevant to the objectives of the study were then used as data for the analysis.

\section{Results}
This section provides the results of the pilot study conducted to constitute an initial test of the collaborative interaction setup. The first section gives an overview of robot language acquisition. The second section presents the results of the participants' experience as obtained from the survey and semi-structured interview thematic analysis. 

\subsection{Robot Language Acquisition}

The robot's ability to learn to associate certain words in its vocabulary with a certain internal need is determined by convergence rate (as described in 2.2). Figure 3 visualises the mean convergence rate for DOT (left) and non-DOT (right) conditions, respectively. No condition met full convergence, possibly due to the limited number of iterations. However, in the DOT condition a total higher average reward was achieved than in the non-DOT condition.

\begin{figure}
    \centering
\includegraphics[width=0.8\textwidth]{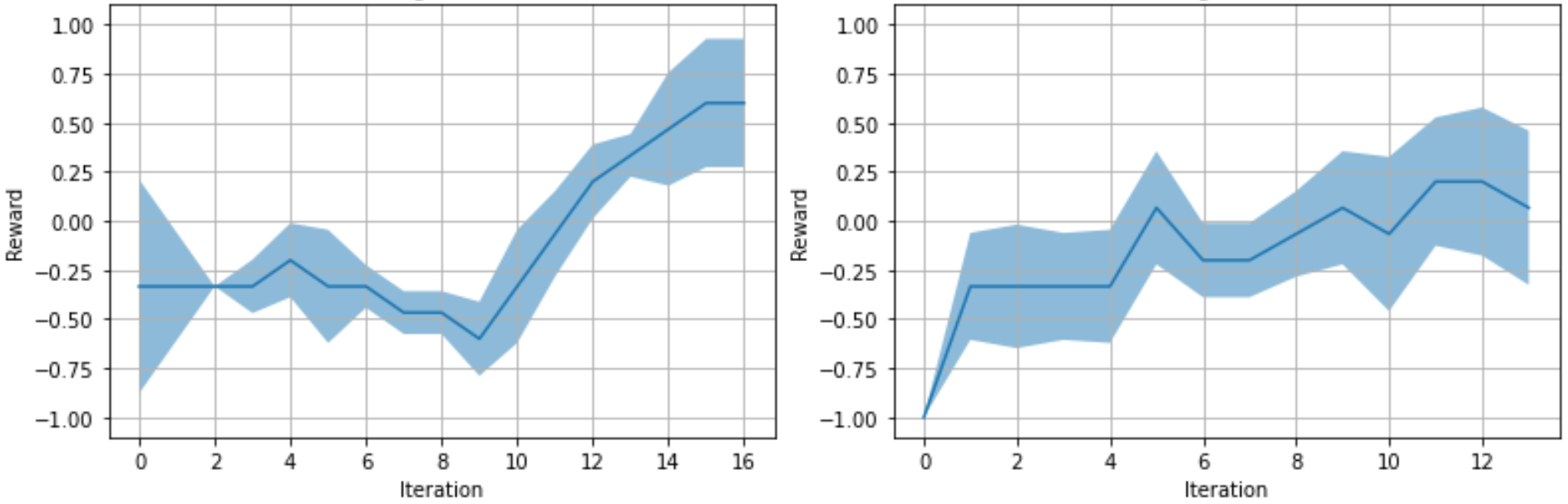}
\caption{Average reward and standard deviation obtained by the robot in the DOT (left) and non-DOT (right) condition over the mean number of iterations. Iterations ended automatically as a function of learning.} \label{fig3}
\end{figure}

\subsection{Participant Experience}
The mean self-reported affective ratings of the SAM-scale are displayed in Figure 4. Participants generally reported high valence (happy) (M=3.5) and dominance (feeling of being in control) (M=3.7) as well as low arousal (not stressed) (M=2.5). The results of the additional questions about the participants' experiences can be seen in Table 1. 

Furthermore, a thematic analysis of the semi-structured interviews revealed some insights into the participants' experience of interacting with the robot and the setup. Most participants found the task to be easy to understand, although lacking in being grounded in a sufficient purpose. Most participants found it easy to differentiate whether the robot was given an object it wanted or not. Commonly occurring themes were \textbf{confusion} but at the same time \textbf{amusement} and \textbf{excitement}. Another theme was \textbf{negative/positive coding} of the robot's movements and sounds (as expressed as audiovisual feedback); most participants found the antennae to be natural and intuitive to interpret as positive or negative. However, a commonly occurring theme regarding the robot's arm movement was \textbf{uncanniness/unnaturalness}. There also seemed to be a general confusion about whether certain reactions were stronger than others (e.g. if arm movements meant \textit{more} happy than e.g. head movements). This was expressed by a participant saying ``\textit{The logic of the antennas made a lot of sense. But I did not understand the intensity of the body language, if certain movements were more positive than others.}" Another participant stated that ``\textit{...it has too large arms and it was unsettling, but the rest of the movements were pretty cute. The antennae felt like ears and since animals are familiar when they go up or down for happy and sad the antennae worked the same way.}" Another common theme concerned the robot's \textbf{age} and \textbf{child-likeness}. Most participants expressed some form of care or empathy towards the robot, which was generally perceived as a 1-3 year old, evoking feelings of wanting to provide it with its needs. One participant reflected on the child similarities by stating ``\textit{It seems like a real child because it is using trial and error with the words it uses, sees what happens and then express either happiness or disappointment. What is not similar is that it is not trying same word again, a real child would have kept using the same word until it got what it wanted.}" A final theme was \textbf{strategies} to approach the task. A few of the participants revealed trying to use some strategy to make the task easier and for learning, such as exclusion methods or ``programming" the robot by intentionally giving certain objects for certain words. Finally, the simulated environment was generally negatively perceived as unnatural and game-like, making the task feel like a simulation; one participant said ``\textit{The design itself is very cute. It would be different with a real robot, a robot on a screen is more like a game rather than a real interaction.}"

\begin{table}[htbp]
    \centering
    \caption{Mean Likert Scale Survey Results (1: Negative, 5: Positive)}
    \label{tab:survey-results}
    \begin{tabular}{lcc}
        \toprule
        \textbf{Question} & \textbf{Mean (SD)} \\
        \midrule
        How did you experience the robot’s reactions when given an object it wanted? & 4.67 (0.33) \\
        How did you experience the robot’s visual response when given an object it wanted? & 3.67 (1.03) \\
        How did you experience the robot’s audible response when given an object it wanted? & 4.67 (0.33) \\
        How did you experience the robot’s response when given an object it didn't want? & 1.33 (0.52) \\
        What was your impression of the robot's appearance? & 3.50 (1.29) \\
        \bottomrule
    \end{tabular}
\end{table}

\begin{figure}
    \centering
    \includegraphics[width=0.4\textwidth]{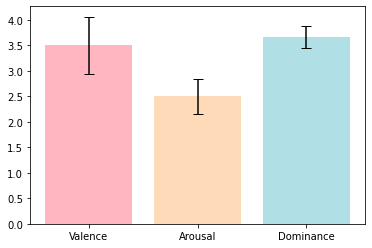}
    \caption{Mean results (and standard error) of the Self Assessment Manikin Scale's three dimensions of self-reported affective ratings. Each dimension is measured between 1 to 5 for valence (5=happy, 1=unhappy), arousal (5=stressed, 1=not stressed) and dominance (5=in control, 1=not in control)}
    \label{fig4}
\end{figure}

\section{Discussion and future work}
This study investigated a human robot mutual learning setup where the robot's ability to learn to associate words with internal needs, was determined by convergence rate and human experience was determined by self-reported affective ratings and a semi-structured interview and thematic analysis. The results, as shown in Figure 3, indicate that neither in the DOT nor the non-DOT condition did the robot achieve full convergence. It is, however, noteworthy that the DOT condition demonstrated a higher rate of learning compared to the non-DOT condition. This suggests that the robot's learning process was influenced by the human's learning being enhanced by the DOT, but further exploration with more iterations as well as more participants is necessary for conclusive results. Considering the low number of participants in this study, it would be of interest for further studies to assess whether the mean differences between DOT and non-DOT show a statistical significance in terms of both human correct responses (to the robot's babblings) and robot language convergence rate. Investigating this DOT effect in the participants could provide further insights into how effective a setup like this could be for cognitive interventions, including memory training for people with dementia and MCI. Furthermore, the language acquisition model in this study is still in its infancy and thus quite simple and limited to a few internal needs. Future work will expand this model to better capture language learning and interaction complexities, making the model more generalizable and scalable, the current study is thus an important step in that direction. 

The affective ratings, displayed in Figure 4, showed that participants reported high valence (happiness) and dominance (being in control), along with low arousal (not feeling stressed). The thematic analysis of the semi-structured interviews showed an emphasis on attitudes towards the setup, where some found the task relatively easy to comprehend, but others expressed a desire for a more meaningful and less simulated purpose behind the interaction. This suggests that incorporating a clearer objective or validation with the physical robot could enhance the user experience. Another theme was the negative and positive coding of the robot's feedback and the results from the survey suggested that the positive audio feedback was more indicative of a positive response than the visual (movement) feedback. Fine-tuning the robot's various signals could improve the user's understanding of its feedback and thus also further enhance the setup, the DOT and its efficiency. Furthermore, the perception of the robot itself, including age, child-likeness and uncanniness were commonly occurring themes. The question of whether the robot being child-like facilitated the interaction or not remains to be investigated in future studies. A large amount of the socially assistive robots used for elderly care in particular tends to be designed to resemble either a pet or a child that needs to be cared for \cite{martinez2018personal}. However, in the context of cognitive training and interventions the effectiveness of robot design in terms of age, anthropomorphism and social and affective role requires further investigation, e.g. through assessment of dimensions of warmth and competence \cite{carpinella2017robotic}. Some participants attempted to strategize the robot's learning process by using exclusion methods and intentionally providing certain objects for specific words. This finding demonstrates participants' active engagement in the interaction and highlights the potential for further exploring different human-robot collaborative learning approaches/strategies. 

In conclusion, the pilot study offers some promising results for the mutual learning human-robot interaction setup as proposed in this paper. The findings tentatively suggest that incorporating DOT could enhance both the human and the robot's learning process. Moreover, participants generally had positive emotional experiences and expressed care and empathy towards the robot. To build upon these findings, future studies should consider increasing the number of iterations in the language acquisition process as well as the number of participants to evaluate the convergence rate and DOT-effect more comprehensively. Additionally, providing a clearer purpose or context to the interaction and refining the robot's audiovisual feedback could lead to a more immersive and meaningful experience for users. Furthermore, investigating how users' strategies and actions during the interaction impact the robot's learning process could be of interest for collaborative human-robot learning systems. Finally, exploring interactions in more ecologically valid, i.e. physical environments settings with complex homeostatic drives, rather than a simulated environment and a robot with relatively simple homeostatic drives, may offer a more authentic and comprehensive understanding of human-robot interaction based on mutual learning with respect to homeostatic and affective robotic states. Ultimately, a target group for this work is persons with Mild Cognitive Impairment or dementia undergoing cognitive training based interventions. Given further study and development of our approach, e.g. to include more complex drives and use of the physical robot, the mutual learning for a robot with homeostatic needs potentially provides: a) social (e.g. nurturing), b) `active' (or co-) learning benefits to such participants that might provide positive therapeutic outcomes beyond more standard `passive' learning practises.  

%
%

\bibliographystyle{splncs04}

\bibliography{references}

\end{document}